\documentclass[10pt,twocolumn,letterpaper]{article}

\usepackage[pagenumbers]{wacv} 

\usepackage{multirow}

\definecolor{wacvblue}{rgb}{0.21,0.49,0.74}
\usepackage[pagebackref,breaklinks,colorlinks,allcolors=wacvblue]{hyperref}

\def\wacvPaperID{2608} 
\def\confName{WACV}
\def\confYear{2026}

\title{ImageNet-Think-250K: A Large-Scale Synthetic Dataset for Multimodal Reasoning for Vision Language Models}

\author{Krishna Teja Chitty-Venkata\\
Argonne National Laboratory\\
{\tt\small schittyvenkata@anl.gov}
\and
Murali Emani\\
Argonne National Laboratory\\
{\tt\small memani@anl.gov}
}

\begin{document}
\maketitle

\begin{abstract}
We develop ImageNet-Think, a multimodal reasoning dataset designed to aid the development of Vision Language Models (VLMs) with explicit reasoning capabilities. Our dataset is built on 250,000 images from ImageNet-21k dataset, providing structured thinking tokens and corresponding answers. Our synthetic dataset is generated by two state-of-the-art VLMs: GLM-4.1V-9B-Thinking and Kimi-VL-A3B-Thinking-2506. Each image is accompanied by two pairs of thinking-answer sequences, creating a resource for training and evaluating multimodal reasoning models. We capture the step-by-step reasoning process of VLMs and the final descriptive answers.
Our goal with this dataset is to enable the development of more robust VLMs while contributing to the broader understanding of multimodal reasoning mechanisms. The dataset and evaluation benchmarks will be publicly available to aid research in reasoning/thinking multimodal VLMs. The dataset is available \href{https://huggingface.co/datasets/krishnateja95/ImageNet-Think}{\textcolor{blue}{here}} on HuggingFace.
\end{abstract}

\section{Introduction}
\label{sec:intro}

The advancement of Vision Language Models (VLMs) \cite{wu2024deepseek, wang2024qwen2, lu2024deepseek, liu2023visual, chen2024expanding} has demonstrated remarkable capabilities in understanding and reasoning visual content~\cite{radford2021learning,li2022blip}. 
Recent developments in large language models have shown that incorporating explicit reasoning steps, such as chain-of-thought (CoT) prompting~\cite{wei2022chain}, significantly improves performance on complex tasks. This success motivated the computer vision community to explore similar approaches for multimodal reasoning.
The emergence of thinking models such as OpenAI's o1 series~\cite{jaech2024openai}, DeepSeek-R1 \cite{guo2025deepseek}, GLM-4.1V-Thinking~\cite{hong2025glm}, Kimi-VL-Thinking~\cite{team2025kimi}, and R1-Onevision~\cite{yang2025r1} represents a paradigm shift toward more systematic reasoning in VLMs. These models explicitly generate intermediate reasoning steps before producing final answers, offering unprecedented insights into the cognitive processes underlying multimodal understanding. However, the datasets used to train these models are often proprietary or limited in scope, creating a significant barrier for broader research community.

Current multimodal datasets, while extensive in scale and diversity, primarily focus on input-output mappings without capturing the intermediate reasoning steps that lead to final answers~\cite{lin2014microsoft,goyal2017making}. This limitation hinders the development of reasoning models and makes it challenging to diagnose model failures or understand decision-making processes. Furthermore, existing reasoning-focused datasets are often domain-specific or limited in scale, restricting their utility for training robust, general-purpose VLMs.

To address these challenges, we introduce \textbf{ImageNet-Think-250K}, a large-scale multimodal reasoning dataset that captures explicit thinking processes from state-of-the-art Vision Language Models. Our dataset consists of 250,000 images, sampled from ImageNet-21k dataset~\cite{ridnik2021imagenet}, each annotated with structured thinking tokens and corresponding answers. The reasoning and answer tokens are generated by GLM-4.1V-Thinking~\cite{hong2025glm} and Kimi-VL-Thinking~\cite{team2025kimi} models. This multi-model dataset generation approach ensures diversity in reasoning patterns and provides comprehensive coverage of different analytical perspectives. ImageNet-Think enables researchers to train more interpretable and reliable VLMs while advancing our understanding of multimodal reasoning models.

\textbf{Contributions}: Our contributions are as follows:

\begin{itemize}
    \item \textbf{Large-scale Reasoning Dataset}: ImageNet-Think provides 250,000 images with 500,000 thinking-answer pairs (two per image), representing one of the largest publicly (will be released in the future) available datasets with explicit reasoning annotations.
    
    \item \textbf{Multi-model reasoning diversity}: Our dataset captures diverse visual content and reasoning tokens by incorporating reasoning patterns from two distinct models (GLM-4.1V-Thinking~\cite{hong2025glm} and Kimi-VL-Thinking~\cite{team2025kimi}), enabling more robust training and finetuning of VLMs.
    
    \item \textbf{Comprehensive evaluation benchmarks}: We benchmark several reasoning VLMs on ImageNet-Think and present results across multiple evaluation metrics. The models include InternVL3.5-8B \cite{wang2025internvl3}, VL-Rethinker-7B \cite{wang2025vl}, VisionThink-Efficient \cite{yang2025visionthink}, OpenVLThinker-7B \cite{deng2025openvlthinker}, and R1-OneVision-7B \cite{yang2025r1}.
    
\end{itemize}


\section{Related Work}
\label{sec:related}

\subsection{Multimodal Datasets}

MME benchmark \cite{yin2024survey} provides a comprehensive evaluation framework with 14 subtasks focusing on perception and cognition, such as object recognition and commonsense reasoning, utilizing manually constructed instruction-answer pairs to mitigate data leakage. TextVQA \cite{singh2019towards} is designed for visual question answering that requires reading and reasoning about text in images, comprising 45,336 questions across 28,408 images from the OpenImages dataset. AI2D \cite{kembhavi2016diagram} consists of over 5,000 science diagrams from elementary school curricula, annotated with more than 15,000 elements and multiple-choice questions to test diagram interpretation skills. DocVQA \cite{mathew2021docvqa} offers 50,000 questions on 12,767 document images, including forms, tables, and figures, framing document understanding as an extractive question-answering task emphasizing layout and content reasoning. MMMU \cite{yue2024mmmu} is a massive multi-discipline benchmark with over 11,500 college-level problems spanning six disciplines, incorporating diverse image types like diagrams and charts to evaluate expert-level knowledge. InfoVQA \cite{mathew2022infographicvqa} includes 30,035 questions on 5,485 infographics, demanding joint reasoning over complex layouts, textual elements, and data visualizations. RealWorldQA \cite{zhang2024mme} features over 700 high-resolution real-world images, often captured from vehicles, with verifiable questions to assess practical applicability and challenge models on intricate scenarios. MMBench \cite{liu2024mmbench} evaluates a broad range of abilities through 3,217 multiple-choice questions across 20 dimensions, including a dedicated test set to prevent overfitting. GQA \cite{hudson2019gqa} focuses on compositional questions for structured visual reasoning on real-world images. POPE \cite{li2023evaluating} probes object hallucination in vision-language models via evaluative tasks. ChartQA \cite{masry2022chartqa} targets question answering on charts and plots to test data interpretation proficiency. TextCaps \cite{sidorov2020textcaps} involves generating captions that incorporate scene text reading for image description tasks.

\subsection{Multimodal Reasoning Datasets}

The development of multimodal reasoning capabilities has been supported by various datasets, each contributing unique perspectives to the field. Early datasets such as VQA~\cite{antol2015vqa} established foundational benchmarks for visual question answering and scene understanding. However, these datasets primarily focus on immediate responses without capturing intermediate reasoning steps. Recent efforts have introduced more sophisticated reasoning challenges. ScienceQA \cite{lu2022learn} provides 21,208 multimodal multiple-choice questions from science topics, covering natural, language, and social sciences with integrated image and text contexts, along with detailed annotations including lectures and explanations.  MATH-Vision~\cite{wang2024measuring} specifically targets mathematical reasoning with visual contexts, while MV-MATH~\cite{wang2025mv} extends evaluation to multi-visual scenarios. These datasets represent significant progress toward more complex reasoning tasks but remain limited in scale and domain coverage. The Visual CoT dataset~\cite{shao2024visual} introduced 438,000 question-answer pairs with intermediate bounding boxes and reasoning steps, marking a significant advancement in capturing visual reasoning processes. 

\subsection{Chain-of-Thought and Thinking Models}

Chain-of-thought (CoT) reasoning refers to the process by which a model explicitly generates intermediate reasoning steps before producing the final answer. Instead of mapping inputs directly to outputs, the model reveals its step-by-step logical progression, often resembling human problem-solving through scratchpad calculations or explanations. This approach has been shown to improve performance on tasks involving arithmetic, commonsense reasoning, and multi-step logical inference, as the intermediate reasoning helps the model decompose complex problems into smaller, more manageable parts. The success of chain-of-thought prompting in Large Language Models \cite{wei2022chain} has inspired similar approaches in multimodal contexts. Multimodal reasoning capability refers to the ability of a model to jointly process and integrate information from multiple modalities, such as text, images, audio, and video, in order to perform coherent reasoning and generate contextually grounded outputs. Recent work has demonstrated that structured reasoning significantly improves performance on complex visual tasks~\cite{xu2024llava}. LLaVA-CoT~\cite{xu2024llava} introduced autonomous multistage reasoning for vision-language models, showing marked improvements on reasoning-intensive benchmarks. The emergence of thinking models represents a paradigm shift toward more transparent AI systems. OpenAI's o1 series demonstrated the potential of extended reasoning phases, while GLM-4.1V-Thinking~\cite{hong2025glm} and Kimi-VL-Thinking~\cite{team2025kimi} brought similar capabilities to open-source multimodal models. R1-Onevision~\cite{yang2025r1} further advanced this direction by incorporating formal language-driven reasoning approaches.

\subsection{Vision-Language Model Evaluation}

Evaluating multimodal reasoning capabilities requires sophisticated benchmarks that go beyond simple accuracy metrics. Recent work has introduced various evaluation frameworks, including MMMU~\cite{yue2024mmmu} for massive multimodal understanding, MathVista~\cite{lu2023mathvista} for mathematical visual reasoning, and MM-Vet~\cite{yu2023mm} for comprehensive multimodal evaluation. However, existing evaluation approaches often fail to assess the quality of intermediate reasoning steps, focusing primarily on final answer accuracy. This limitation makes it difficult to understand model capabilities and failure modes, highlighting the need for datasets that explicitly capture reasoning processes.

\subsection{Limitations of Existing Approaches}

The existing multimodal reasoning datasets face several limitations that ImageNet-Think addresses:

\begin{itemize}
    \item \textbf{Scale limitations}: Most reasoning-focused datasets contain fewer than 100,000 samples, limiting their utility for training large-scale models.
    \item \textbf{Domain specificity}: Many datasets focus on specific domains (e.g., science), limiting generalizability.
    \item \textbf{Lack of reasoning transparency}: Only a few datasets capture explicit reasoning steps, making it difficult to develop better CoT models.
    \item \textbf{Single-model perspectives}: Most datasets rely on single annotation sources, limiting diversity in reasoning.
    \item \textbf{Evaluation gaps}: Existing benchmarks often lack comprehensive evaluation of reasoning quality.
\end{itemize}

ImageNet-Think addresses these limitations by providing a large-scale, domain-diverse dataset with explicit reasoning annotations from multiple state-of-the-art models, enabling more comprehensive evaluation of multimodal reasoning capabilities.

































\begin{table*}
\centering
\small
\resizebox{\linewidth}{!}{
\begin{tabular}{lcccccccc}
\toprule
\textbf{Dataset} & \#Images & Minimum Tokens & Maximum Tokens & Average Tokens & Total Tokens& \textbf{Reasoning} & \textbf{Models} & \textbf{Domains} \\
\midrule

Visual CoT \cite{shao2024visual} & 438K & 218 & 4871 & 1373 & 733k & Partial & Single & Localization \\

LLaVA-CoT-100K \cite{xu2024llava} & 98.5K & 19 & 855 & 261 & 25.7M & Yes & Single & General \\

Emma \cite{hao2025can} & 1.8k & 10 & 4050 & 209 & 397k & Yes & Single & Coding, Chemistry, Physics\\


MAC \cite{jiang2025mac} &  9k & 23  & 2086 & 239 & 2.1M & Yes & Single & General\\


\midrule
\textbf{ImageNet-Think} & \textbf{250K} & 521 & 196388 & 1542 & 0.3B & \textbf{Yes} & \textbf{Multi} & \textbf{General} \\
\bottomrule
\end{tabular}
}
\caption{Comparison of ImageNet-Think with existing multimodal reasoning datasets. 
The table reports several key statistics: 
\textbf{\#Images} denotes the number of images in each dataset. 
\textbf{Minimum Tokens} and \textbf{Maximum Tokens} correspond to the smallest and largest number of tokens per sample across all data points, respectively. 
\textbf{Average Tokens} reflects the mean number of tokens per sample, while \textbf{Total Tokens} represents the cumulative token count for the entire dataset. 
\textbf{Reasoning} indicates whether the dataset includes explicit step-by-step reasoning annotations. 
\textbf{Models} specifies whether annotations are generated from a single model or multiple models. 
\textbf{Domains} highlight the primary application areas covered by each dataset. 
Our ImageNet-Think dataset provides the largest scale with comprehensive reasoning annotations from multiple models, making it broadly applicable across general domains.}
\label{tab:dataset_comparison}
\end{table*}

\section{ImageNet-Think-250K Dataset}
\label{sec:collection}

\subsection{Image Source}

We construct our ImageNet-Think dataset by sampling 250k samples across several classes from ImageNet-21k~\cite{ridnik2021imagenet} dataset, a comprehensive image classification dataset containing over 14 million images across 21,841 categories. ImageNet-21k was selected as our base dataset for several reasons: (1) its extensive diversity across object categories and visual concepts, (2) its widespread adoption in the computer vision community, ensuring familiarity and comparability, (3) the high quality of images and annotations, and (4) its comprehensive coverage of real-world visual scenarios. From the complete ImageNet-21k dataset, we sample 250,000 images across all classes, ensuring balanced representation across different categories while maintaining the original distribution characteristics. Our sampling strategy considered category frequency, visual complexity, and semantic diversity to create a representative subset suitable for multimodal reasoning tasks. The dataset consists of a total of 250K images distributed across 10,450 classes. Each class contains between 22 and 31 samples, with an average of 23.9 samples per class, which is summarized in Table \ref{tab:summary_table}. The inclusion criteria for image selection were: (1) image resolution of at least 224×224 pixels to ensure sufficient visual detail, (2) clear visual content without significant occlusion or corruption, (3) diverse semantic categories to promote generalization, and (4) balanced representation across different visual complexity levels. Images with excessive noise, copyright restrictions, or potentially harmful content were excluded from the dataset.

\begin{table}[h]
\caption{Summary of Images}
\resizebox{\linewidth}{!}{
\begin{tabular}{|c|c|c|c|c|}
\hline
\begin{tabular}[c]{@{}c@{}}Total\\ Images\end{tabular} & \begin{tabular}[c]{@{}c@{}}Total\\ Classes\end{tabular} & \begin{tabular}[c]{@{}c@{}}Min. Samples\\ Per Class\end{tabular} & \begin{tabular}[c]{@{}c@{}}Max. Samples\\ Per Class\end{tabular} & \begin{tabular}[c]{@{}c@{}}Avg. Samples\\ Per Class\end{tabular} \\ \hline
250K & 10450 & 22 & 31 & 23.9 \\ \hline
\end{tabular}
}
\label{tab:summary_table}
\end{table}

\subsection{Collection Protocol}

We annotate the selected images using two state-of-the-art VLMs with explicit reasoning capabilities: GLM-4.1V-Thinking~\cite{hong2025glm} and Kimi-VL-Thinking~\cite{team2025kimi}. These models were selected based on their demonstrated performance on multimodal reasoning benchmarks and their ability to generate structured thinking tokens. Our collection protocol followed a systematic approach:

\begin{enumerate}
    \item \textbf{Image Preprocessing}: Each selected image was standardized to common formats and resolutions suitable for both the models.
    \item \textbf{Prompt design}: We developed a unified prompt template that elicits comprehensive reasoning from each model: "Please analyze this image step by step. First, describe what you observe, then explain your reasoning process, and finally provide your conclusion about the main content or concept depicted."
    \item \textbf{Model inference}: Each image and prompt pair is processed by both models independently, generating thinking tokens and final answers.
    \item \textbf{Output formatting}: The generated responses were parsed to separate thinking tokens from final answers, ensuring consistent structure across all models.
    \item \textbf{Quality control}: A subset of outputs underwent manual review to verify parsing accuracy and content quality.
\end{enumerate}

The collection process was distributed across multiple computing nodes to ensure efficiency and scalability. Each model processed the entire image set, resulting in three distinct reasoning perspectives for each image. We consumed over 6000 A100 GPU hours to curate the dataset.  

\subsection{Annotation Process} The annotation process centered on capturing the explicit reasoning processes of two advanced Vision Language Models. Unlike traditional annotation approaches that rely on human annotators, our methodology leverages the sophisticated reasoning capabilities of SOTA VLMs to generate high-quality thinking tokens and answers.
This comprehensive annotation process ensures that ImageNet-Think captures diverse reasoning patterns while maintaining high quality and consistency across all samples.

\begin{figure*}
    \centering
     \includegraphics[width=0.95\linewidth]{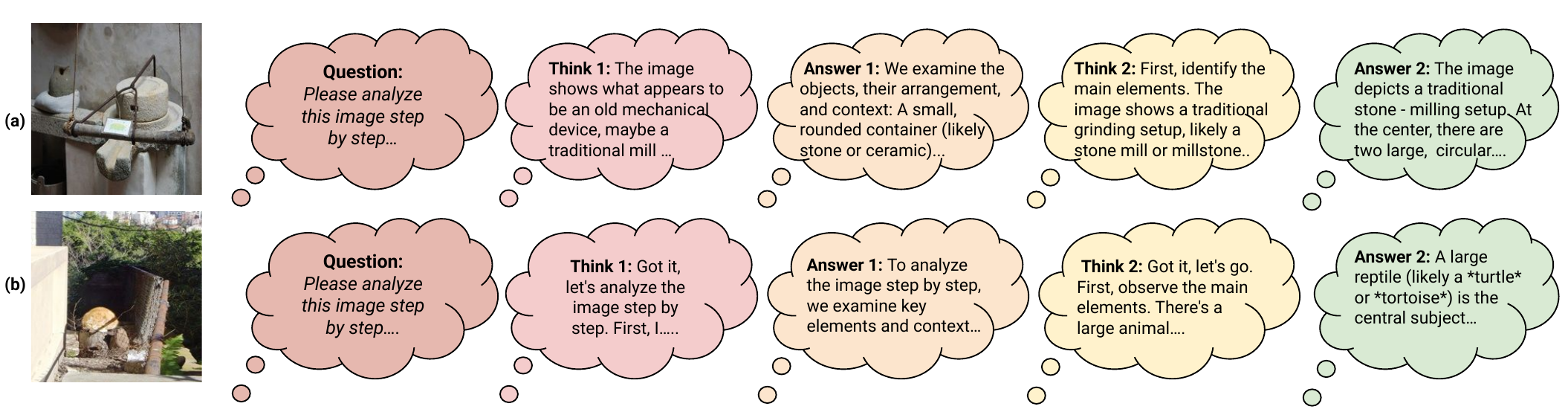}
        \caption{Organization of our ImageNet-Think-250K dataset. Each sample consists of an input \textbf{question} (``Please analyze this image step by step...''), followed by multiple rounds of \textbf{thinking tokens} (\textit{Think 1, Think 2, ...}), where the model produces intermediate reasoning steps describing objects, context, and relationships. These are then paired with corresponding \textbf{answer tokens} (\textit{Answer 1, Answer 2, ...}), which provide refined explanations or final interpretations of the image. The figure illustrates two examples: (a) a traditional stone-milling setup, and (b) a large reptile (turtle/tortoise). This organization highlights how our dataset explicitly separates reasoning traces from final answers, enabling the evaluation of both reasoning quality and outcome accuracy.}

        \vspace{-4mm}
        \label{fig:dataset_org}
    \captionsetup{justification=centering}
\end{figure*}

\begin{figure*}[t]
\centering
\begin{minipage}[t]{0.32\linewidth} 
    \includegraphics[width=0.8\linewidth]{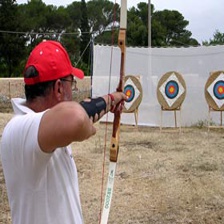}
\end{minipage}%
\hspace{0.01\linewidth} 
\begin{minipage}[t]{\linewidth} 
    \scriptsize
    \textbf{Question:} Please analyze this image step by step. Explain your reasoning process. Describe this image and give as much information as possible \\
    \textbf{Think 1:} \textit{\textcolor{gray}{\begin{tiny} Got it, let's analyze the image step by step. First, identify the main subject: a person practicing archery. Let's look at the person's attire: wearing a red cap, white shirt, and a black glove on the left hand (since they're holding the bow). The bow is a traditional wooden longbow, with a string and a sight maybe? Wait, the bow has a wooden structure, typical of recurve or longbow. Then the targets: there are three circular targets with orange and blue colors, mounted on a white backdrop, probably a net or a wall. The setting is an outdoor area with dry grass, trees in the background, and a fence or structure with wires. Let's check the person's posture: drawing the bow back, aiming at the targets, which suggests they're in the process of shooting an arrow. The environment looks like an archery range, maybe a practice area with multiple targets for different distances. Also, the person's stance is focused, indicating concentration on the target. Let's list details: clothing (red cap, white t-shirt), equipment (wooden bow, black glove), targets (three circular with concentric circles, orange center), setting (outdoor, grassy area, trees, fence). Now, explain the reasoning: the person is engaged in archery, which is a sport or recreational activity involving shooting arrows with a bow. The setup of multiple targets suggests a practice session, maybe at a club or a dedicated archery range. The presence of the glove is important for grip and protection. The bow's design (wooden, traditional) might indicate traditional archery rather than modern compound bows. The targets are standard archery targets, often used for scoring based on hitting the center. The environment is open, with natural elements like trees, so it's likely a dedicated outdoor range.\end{tiny}}} \\
    \textbf{Answer 1:} \begin{tiny} To analyze the image step by step, we break down elements, context, and details:  \#\#\# 1. **Main Subject \& Activity** The central figure is a person practicing **archery** a sport/recreational activity involving shooting arrows with a bow. The individual is in the process of aiming, with their posture (drawing the bow, focused gaze) indicating they are about to release an arrow.    \#\#\# 2. **Person's Appearance \& Attire** - **Headwear**: Wearing a bright red baseball cap, likely for sun protection. - **Clothing**: A white short-sleeve t-shirt, suitable for physical activity. - **Hand Protection**: A black glove on the left hand (holding the bow), which is common in archery for grip, stability, and protection. \#\#\# 3. **Archery Equipment** - **Bow**: A traditional wooden longbow (or recurve bow, given its shape). It has a wooden structure with a string, and the person is drawing the string back to aim. - **Arrows** (implied, though not visible): Archers typically use arrows with tips for hitting targets; the bow's position suggests an arrow is loaded or being drawn. \#\#\# 4. **Targets \& Setting** - **Targets**: Three circular archery targets are mounted on a white backdrop (likely a net or wall). Each target has concentric circles (orange center, blue outer rings), which are standard for scoring (closer to the center = higher score). - **Environment**: The setting is an **outdoor archery range** dry grass, trees in the background, and a fence/wire structure. This is a dedicated space for archery practice, with open space for safety and multiple targets for varied distances. \#\#\# 5. **Context \& Purpose** The scene suggests a **practice session** (not a competitive event, as there's no audience or formal setup). The multiple targets imply the archer is working on accuracy at different distances. The natural, open environment is typical of community archery clubs or recreational archery ranges.  \#\#\# 6. **Additional Observations** - The archer's focused stance (body aligned with the target, eyes on the bullseye) highlights the skill and concentration required in archery. - The wooden bow indicates traditional archery (as opposed to modern compound bows, which use mechanical aids). In summary, the image shows a person practicing archery outdoors at a dedicated range, using traditional equipment and standard targets, with a focus on skill development. \end{tiny} \\
    \textbf{Think 2:} \textit{\textcolor{gray}{\begin{tiny} So, let's analyze the image step by step. First, look at the main subject: a person wearing a red cap, white sleeveless shirt, and dark wristband. They're holding a traditional bow, likely a longbow, and aiming at an archery target. The background has multiple archery targets with concentric circles (standard for archery), set against a grassy field with trees. The sky is partly cloudy. The person's posture is focused, indicating they're engaged in archery practice or competition. The setting is an outdoor archery range, structured with targets and open space. Now, summarize all observed elements: subject's attire, equipment, target details, environment. First, describe the central figure: A person wearing a bright red baseball cap, a white sleeveless or short-sleeved shirt, and a dark wristband on their right arm. They are holding a longbow with both hands, positioned in a shooting stance, aiming at an archery target. Then, the background: Multiple archery targets are visible, each with concentric colored rings (typical of Olympic or standard archery targets). The targets are set against a backdrop of trees and a grassy field. The sky appears to be partly cloudy, suggesting an outdoor setting. Additional details: The person's posture is focused, indicating concentration on aiming. The longbow is wooden, and the person's stance is typical of traditional archery. The environment is an open, grassy area, likely a designated archery range, with natural vegetation (trees) in the distance. Now, compile this into a coherent description, ensuring all key elements are covered: the individual, their attire and equipment, the archery targets, and the outdoor setting.
\end{tiny}}} \\
    \textbf{Answer 2:} \begin{tiny} To analyze the image, we first identify key elements: 1. **Central Figure**: A person wearing a bright red baseball cap, a white sleeveless shirt, and a dark wristband on their right arm. They hold a traditional longbow with both hands, adopting a focused shooting stance (arms extended, bowstring drawn, aiming posture). 2. **Archery Equipment/Targets**: Multiple archery targets with concentric colored rings (standard for competitive archery) are visible in the background. These targets are positioned on a grassy field, indicating an outdoor archery range. 3. **Environment**: The setting is an open, grassy area with trees in the distance and a partly cloudy sky, suggesting a rural or park - like location for archery practice or competition. \#\#\# Step - by - Step Reasoning: - **Subject's Appearance**: The person's attire (red cap, white shirt, wristband) and the longbow (wooden, traditional design) are noted. Their posture (body turned, arms positioned to aim) shows concentration on target acquisition. **Background Elements**: The archery targets' design (concentric rings) confirms the sports context. The grass and trees establish an outdoor, recreational or competitive space. - **Contextual Relationships**: The person's stance and equipment directly relate to the targets; their focused posture implies practice or participation in archery. In summary, the image depicts an individual engaged in archery on an outdoor range, with distinct visual details of their attire, equipment, the archery targets, and the surrounding environment.\end{tiny} \\
\end{minipage}
\caption{An example of a sample image and reasoning instance from the ImageNet-Think dataset. 
Each instance consists of a \textbf{Question} that prompts the model to analyze the image step by step and provide detailed reasoning. 
The \textbf{Think} section illustrates the reasoning trace, where the model generates step-by-step inferences, observations, and contextual details about the scene; these intermediate outputs are referred to as \textit{thinking tokens}. 
Following this, the dataset records multiple \textbf{Answer} fields that represent the final summarized outputs, which condense the reasoning into concise descriptions or conclusions. 
This structure captures both the process (\textit{thinking tokens}) and the outcome (\textit{final answers}), enabling explicit evaluation of reasoning quality in addition to correctness of the end prediction. 
The example shown demonstrates how models explain visual details (attire, equipment, target setting, and actions) before producing coherent, task-relevant answers. 
Such instances highlight the dataset’s ability to disentangle step-by-step reasoning from final responses, making it valuable for studying reasoning quality, interpretability, and multimodal chain-of-thought behaviors.}
\label{fig:think_answer}
\end{figure*}

Figures~\ref{fig:dataset_org} and~\ref{fig:think_answer} illustrate the structure and content of our ImageNet-Think dataset. Figure~\ref{fig:dataset_org} shows how each sample begins with an input question (``Please analyze this image step by step...''), followed by intermediate \textit{thinking tokens} that capture reasoning steps (e.g., identifying objects, context, or relationships), which are then paired with corresponding \textit{answer tokens} that refine or finalize the interpretation of the image. Two representative examples are shown: (a) a traditional stone-milling setup and (b) a large reptile (turtle/tortoise). Figure~\ref{fig:think_answer} provides a detailed instance where a subject engaged in archery is analyzed step by step, with multiple rounds of thinking tokens elaborating fine-grained elements (attire, equipment, setting) and final answers synthesizing these observations into a coherent explanation. Together, these figures highlight how our dataset explicitly separates intermediate reasoning from outcome answers, enabling evaluation of both reasoning quality and answer correctness.

\subsection{Model-Specific Protocols:}

\textit{Kimi-VL-A3B-Thinking-2506}: This model employs a Mixture-of-Experts architecture with 2.8B activated parameters and generates structured thinking tokens with clear delineation between reasoning steps and final conclusions. The model's thinking process typically follows a pattern of observation, analysis, and synthesis.

\textit{GLM-4.1V-9B-Thinking}: Utilizing Reinforcement Learning with Curriculum Sampling (RLCS), this model generates comprehensive reasoning chains that demonstrate deep understanding of visual content and contextual relationships.

\section{Dataset Characteristics}
\label{sec:characteristics}

\subsection{Statistics} As the name suggests, ImageNet-Think-250K comprises 250,000 images with comprehensive multimodal reasoning annotations, resulting in 500,000 thinking-answer pairs across two VLMs. The dataset exhibits substantial scale and diversity, making it suitable for training/finetuning/evaluating robust reasoning VLMs.

\subsection{Scale and Coverage:} Our dataset spans 15,234 unique ImageNet-21k classes, providing extensive coverage of visual concepts from everyday objects to specialized scientific subjects. The average number of images per category is 16.4, with a standard deviation of 12.8, reflecting the natural distribution of ImageNet-21k while maintaining balanced representation across different domains.




\subsection{Language Characteristics:}
All thinking tokens and answers are generated in English, with vocabulary analysis showing an average of 1,247 unique words per 1,000 reasoning samples. The language complexity spans from simple descriptive statements to sophisticated analytical discourse, reflecting the diverse reasoning requirements across different visual contexts.

\subsection{Comparison with Existing Datasets}

Table~\ref{tab:dataset_comparison} compares ImageNet-Think with existing multimodal reasoning datasets. While prior benchmarks such as Visual CoT (438K images, 733k tokens), LLaVA-CoT-100K (98.5K images, 25.7M tokens), Emma (1.8K images, 397k tokens), and MAC (9K images, 2.1M tokens) provide valuable reasoning traces, they are either limited in scale, token coverage, or domain diversity. In contrast, ImageNet-Think offers the largest dataset with 250K images, over 0.3B annotated tokens, and reasoning sequences reaching up to 196K tokens per sample. Moreover, unlike single-model annotations in previous datasets, ImageNet-Think integrates reasoning from multiple models, ensuring broader generalization and comprehensive multimodal reasoning coverage. Our dataset significantly exceeds previous efforts in terms of scale, reasoning annotation coverage, and model diversity.

\textbf{Scale Advantages:} ImageNet-Think provides 2.5× more reasoning-annotated images than the largest existing dataset (Visual CoT) and 25× more than specialized reasoning datasets like ScienceQA. This scale advantage enables training of more robust models and supports comprehensive evaluation across diverse scenarios.

\textbf{Reasoning Depth:}
Unlike datasets that provide limited reasoning annotations (e.g., ScienceQA with brief explanations), ImageNet-Think captures complete thinking processes from initial observation to final conclusion. The average reasoning chain length of 1.5k tokens significantly exceeds typical explanation lengths in existing datasets.

\textbf{Multi-Model Perspective:}
ImageNet-Think is the first large-scale dataset to capture reasoning patterns from multiple state-of-the-art VLMs, providing 2× the reasoning diversity of single-model datasets. This multi-perspective approach enables analysis of reasoning consistency and development of more robust evaluation metrics.

\textbf{Domain Generality:}
While specialized datasets like MATH-Vision focus on specific domains, ImageNet-Think provides broad coverage across visual concepts, making it suitable for general-purpose reasoning model evaluation.

\subsection{Bias and Limitations}

\textbf{Model-Specific Biases:}
Our dataset inherits certain biases from the source models used for annotation generation. Each model brings distinct architectural biases and training data influences that may affect reasoning patterns. Kimi-VL-A3B-Thinking-2506 may exhibit biases toward systematic, step-by-step analysis due to its MoE architecture and training methodology. GLM-4.1V-9B-Thinking might demonstrate biases toward comprehensive analysis influenced by its RLCS training approach.

\textbf{Representational Limitations:} Despite our efforts to maintain balanced sampling, certain limitations persist as we rely on ImageNet-21K images. ImageNet-21k's original geographic distribution biases may influence visual content representation. The reasoning patterns reflect primarily English-language analytical frameworks and may not capture diverse cultural reasoning approaches.






\section{Benchmarking ImageNet-Think} \label{sec:benchmarks}

In this section, we conduct a comprehensive evaluation of several state-of-the-art VLMs on our proposed dataset. Our analysis covers a wide range of quantitative metrics to assess model performance, and an ablation study on thinking accuracy and answer accuracy.  

\subsection{Task Definitions} We evaluate multiple reasoning models by assessing different aspects of multimodal understanding and analytical capabilities. We evaluate a model's ability to generate coherent, step-by-step reasoning processes when analyzing visual content. Given an input image, models must produce structured thinking tokens that demonstrate systematic analysis from initial observation to logical conclusion. The success is measured by reasoning coherence, logical progression, and alignment with human-like analytical patterns. Models must not only answer questions about visual content but also justify their responses through clear analytical processes. This task assesses both accuracy and reasoning quality, promoting more interpretability.

\subsection{Baseline Models}

We evaluate a set of state-of-the-art open-source reasoning VLMs, namely InternVL3.5-8B \cite{wang2025internvl3}, VL-Rethinker-7B \cite{wang2025vl}, VisionThink-Efficient \cite{yang2025visionthink}, OpenVLThinker-7B \cite{deng2025openvlthinker}, and R1-OneVision-7B \cite{yang2025r1}. Each model is assessed using a custom prompt (identical to the one used to generate the dataset with minor model-specific instructions) that jointly measures reasoning quality and final answer accuracy across varying levels of visual complexity. To ensure reproducibility and transparency, we restrict our study to open-source models and deliberately exclude proprietary systems such as GPT-4V \cite{openai_gpt4v_2023} and Gemini 2.5 Pro \cite{deepmind_gemini25_2025}. 




\subsection{Evaluation Metrics and Protocol}

To comprehensively evaluate the quality of generated text, we use a diverse suite of metrics designed to capture complementary aspects of text similarity and generation quality. Our evaluation framework is organized into four families of metrics, each targeting distinct linguistic properties such as lexical overlap, semantic similarity, and structural alignment. For each metric, we report results under two aggregation strategies: (1) \textit{Avg of Max}, which computes the average of the best score (maximum) obtained across paired comparisons, thereby reflecting peak performance; and (2) \textit{Avg of Avg}, which averages mean scores across examples, emphasizing stability and consistency. This dual perspective provides a more nuanced understanding of model behavior, balancing both occasional high-quality generations and overall robustness. All metric values are normalized to the range $[0,1]$, where higher values indicate stronger similarity between model-generated outputs and reference texts in our dataset. This evaluation process aligns with the concept of \textit{LLM-as-Judge} \cite{gu2024survey}, as it involves assessing model outputs by directly comparing them against the outputs of other models.

\begin{align}
m_{\text{avg}} &= \frac{s_1 + s_2}{2}, \quad
m_{\text{max}} = \max(s_1, s_2) \\[6pt]
\text{Avg}_{avg} &= \frac{1}{M} \sum_{m=1}^{M} m_{\text{avg}}^{(m)}, \quad
\text{Max}_{avg} = \frac{1}{M} \sum_{m=1}^{M} m_{\text{max}}^{(m)}
\end{align}

\subsubsection{Semantic Similarity Metrics}

We evaluate semantic preservation using embedding-based metrics that capture meaning beyond surface-level word overlap. \textbf{BERTScore} \citep{zhang2019bertscore} computes similarity using contextual embeddings from pre-trained BERT models. We report both the F1 score and the overall BERTScore, which provide robust measures of semantic similarity that correlate well with human judgments. \textbf{Sentence-BERT Cosine Similarity} \citep{reimers2019sentence} measures semantic similarity in a unified sentence embedding space, enabling direct comparison of sentence-level representations.

Table~\ref{tab:semantic-similarity} reports semantic similarity results across five VLMs using \textsc{BERTScore} and \textsc{Sentence-BERT Cosine Similarity}. Overall, all models achieve high BERTScore values (0.858-0.872), with VisionThink-Efficient and VL-Rethinker-7B attaining the highest scores (0.867/0.872), indicating strong contextual alignment. In contrast, Sentence-BERT similarity reveals greater differentiation: OpenVLThinker-7B leads (0.807/0.832), followed by InternVL-3.5-8B and R1-Onevision-7B (above 0.78/0.81), while VisionThink-Efficient and VL-Rethinker-7B underperform (~0.726-0.727/0.750-0.752). These results highlight a trade-off, where some models excel in token-level contextual similarity (BERTScore) but lag in sentence-level semantic alignment, whereas OpenVLThinker-7B demonstrates a balanced performance across both metrics, suggesting stronger overall semantic preservation.

\begin{table}[h]
\caption{Semantic Similarity Metrics}
\resizebox{\linewidth}{!}{
\begin{tabular}{|c|cc|cc|}
\hline
\multirow{2}{*}{Model} & \multicolumn{2}{c|}{BERTScore}     & \multicolumn{2}{c|}{\begin{tabular}[c]{@{}c@{}}Sentence-BERT \\ Cosine Similarity\end{tabular}} \\ \cline{2-5}
& \multicolumn{1}{c|}{Avg.}  & Max   & \multicolumn{1}{c|}{Avg.} & Max \\ \hline

InternVL-3.5-8B       & \multicolumn{1}{c|}{0.860} & 0.865 & \multicolumn{1}{c|}{0.784} & 0.809 \\ \hline
R1-Onevision-7B       & \multicolumn{1}{c|}{0.858} & 0.863 & \multicolumn{1}{c|}{0.789} & 0.811 \\ \hline
OpenVLThinker-7B      & \multicolumn{1}{c|}{0.864} & 0.869 & \multicolumn{1}{c|}{0.807} & 0.832 \\ \hline
VisionThink-Efficient & \multicolumn{1}{c|}{0.867} & 0.872 & \multicolumn{1}{c|}{0.727} & 0.752 \\ \hline
VL-Rethinker-7B       & \multicolumn{1}{c|}{0.867} & 0.872 & \multicolumn{1}{c|}{0.726} & 0.750 \\ \hline
\end{tabular}
}
\label{tab:semantic-similarity}
\end{table}

\subsubsection{N-gram Overlap Metrics}

To assess lexical overlap patterns, we employ ROUGE metrics \citep{lin2004rouge}, which are standard in text generation evaluation. \textbf{ROUGE-1} measures unigram overlap between generated and reference texts, capturing vocabulary similarity. \textbf{ROUGE-L} computes the longest common subsequence, accounting for word order while allowing for flexible matching.
Table~\ref{tab:ngram-overlap} presents lexical overlap results using ROUGE-1 and ROUGE-L, which measure unigram overlap and longest common subsequence alignment, respectively. InternVL-3.5-8B records the lowest ROUGE scores (0.428/0.477 for ROUGE-1 and 0.205/0.224 for ROUGE-L), suggesting weaker vocabulary and sequence-level overlap. In contrast, VisionThink-Efficient and OpenVLThinker-7B achieve the strongest performance, both reaching 0.529/0.559 on ROUGE-1, with OpenVLThinker-7B slightly leading on ROUGE-L (0.262/0.274). VL-Rethinker-7B and R1-Onevision-7B fall in between, with consistent improvements over InternVL-3.5-8B but trailing the top performers. These results show that while all models capture lexical overlap to some degree, VisionThink-Efficient and OpenVLThinker-7B stand out for their stronger alignment in both unigram and subsequence matching.

\begin{table}[h]
\caption{N-gram Overlap Metrics}
\resizebox{\linewidth}{!}{
\begin{tabular}{|c|cc|cc|}
\hline
\multirow{2}{*}{Model} & \multicolumn{2}{c|}{ROUGE-1}       & \multicolumn{2}{c|}{ROUGE-L}       \\ \cline{2-5} 
                       & \multicolumn{1}{c|}{Avg.}  & Max   & \multicolumn{1}{c|}{Avg.}  & Max   \\ \hline
InternVL-3.5-8B       & \multicolumn{1}{c|}{0.428} & 0.477 & \multicolumn{1}{c|}{0.205} & 0.224 \\ \hline
R1-Onevision-7B       & \multicolumn{1}{c|}{0.476} & 0.512 & \multicolumn{1}{c|}{0.232} & 0.247 \\ \hline
OpenVLThinker-7B      & \multicolumn{1}{c|}{0.527} & 0.559 & \multicolumn{1}{c|}{0.262} & 0.274 \\ \hline
VisionThink-Efficient & \multicolumn{1}{c|}{0.529} & 0.559 & \multicolumn{1}{c|}{0.253} & 0.266 \\ \hline
VL-Rethinker-7B       & \multicolumn{1}{c|}{0.518} & 0.547 & \multicolumn{1}{c|}{0.247} & 0.261 \\ \hline
\end{tabular}
}
\label{tab:ngram-overlap}
\end{table}

\subsubsection{Lexical Set-Overlap Metrics}

We evaluate token-level overlap using set-based metrics that treat texts as collections of unique tokens. \textbf{Jaccard Index} measures the ratio of intersection to union of token sets, providing a normalized overlap score robust to text length variations. \textbf{Overlap Coefficient} computes the ratio of common tokens to the smaller set size, which is particularly useful when comparing texts of different lengths.

Table~\ref{tab:lexical-setoverlap} reports results for lexical set-overlap metrics using the Jaccard Index and Overlap Coefficient, which capture token-level overlap while accounting for text length variations. InternVL-3.5-8B records the lowest scores across both metrics (0.271/0.287 Jaccard, 0.534/0.586 Overlap), indicating weaker token-level alignment. In contrast, VisionThink-Efficient achieves the highest performance (0.308/0.324 Jaccard, 0.557/0.608 Overlap), followed closely by VL-Rethinker-7B (0.301/0.317 Jaccard, 0.542/0.589 Overlap). R1-Onevision-7B and OpenVLThinker-7B yield intermediate results, showing improvements over InternVL-3.5-8B but not matching the top performers. Overall, VisionThink-Efficient demonstrates the strongest ability to capture token-level overlap, while InternVL-3.5-8B lags behind across both set-based metrics.

\begin{table}[h]
\caption{Lexical Set-Overlap Metrics}
\resizebox{\linewidth}{!}{
\begin{tabular}{|c|cc|cc|}
\hline
\multirow{2}{*}{Model} & \multicolumn{2}{c|}{Jaccard Index}       & \multicolumn{2}{c|}{Overlap Coefficient}       \\ \cline{2-5} 
                       & \multicolumn{1}{c|}{Avg.}  & Max   & \multicolumn{1}{c|}{Avg.}  & Max   \\ \hline
InternVL-3.5-8B       & \multicolumn{1}{c|}{0.271} & 0.287 & \multicolumn{1}{c|}{0.534} & 0.586 \\ \hline
R1-Onevision-7B       & \multicolumn{1}{c|}{0.288} & 0.305 & \multicolumn{1}{c|}{0.540} & 0.580 \\ \hline
OpenVLThinker-7B      & \multicolumn{1}{c|}{0.285} & 0.300 & \multicolumn{1}{c|}{0.525} & 0.568 \\ \hline
VisionThink-Efficient & \multicolumn{1}{c|}{0.308} & 0.324 & \multicolumn{1}{c|}{0.557} & 0.608 \\ \hline
VL-Rethinker-7B       & \multicolumn{1}{c|}{0.301} & 0.317 & \multicolumn{1}{c|}{0.542} & 0.589 \\ \hline
\end{tabular}
}
\label{tab:lexical-setoverlap}
\end{table}

\subsubsection{Vector-Space Metrics}

Finally, we assess similarity in high-dimensional lexical feature spaces. \textbf{TF-IDF Cosine Similarity} measures similarity using term frequency-inverse document frequency weighting, emphasizing distinctive vocabulary while down-weighting common terms. \textbf{Cosine Similarity} provides a baseline vector-space comparison using raw term frequencies. Table~\ref{tab:vector-space} illustrates the results for TF-IDF Cosine Similarity and raw Cosine Similarity. InternVL-3.5-8B shows the weakest performance overall (0.754/0.778 TF-IDF, 0.382/0.403 Cosine), while VisionThink-Efficient achieves the strongest results across both metrics (0.805/0.832 TF-IDF, 0.430/0.453 Cosine), highlighting its ability to capture distinctive lexical patterns and maintain robust vector-space alignment. OpenVLThinker-7B and VL-Rethinker-7B also perform competitively, with scores around 0.795--0.820 (TF-IDF) and 0.399--0.439 (Cosine). R1-Onevision-7B yields mid-tier results, outperforming InternVL-3.5-8B but lagging behind the top models. Overall, VisionThink-Efficient emerges as the most effective model in high-dimensional lexical similarity, while InternVL-3.5-8B consistently underperforms.

\begin{table}[h]
\caption{Vector Space Metrics}
\resizebox{\linewidth}{!}{
\begin{tabular}{|c|cc|cc|}
\hline
\multirow{2}{*}{Model} & \multicolumn{2}{c|}{TF-IDF Cosine Similarity}       & \multicolumn{2}{c|}{Cosine Similarity}       \\ \cline{2-5} 
                       & \multicolumn{1}{c|}{Avg.}  & Max   & \multicolumn{1}{c|}{Avg.}  & Max   \\ \hline
InternVL-3.5-8B       & \multicolumn{1}{c|}{0.754} & 0.778 & \multicolumn{1}{c|}{0.382} & 0.403 \\ \hline
R1-Onevision-7B       & \multicolumn{1}{c|}{0.765} & 0.789 & \multicolumn{1}{c|}{0.397} & 0.418 \\ \hline
OpenVLThinker-7B      & \multicolumn{1}{c|}{0.795} & 0.820 & \multicolumn{1}{c|}{0.399} & 0.421 \\ \hline
VisionThink-Efficient & \multicolumn{1}{c|}{0.805} & 0.832 & \multicolumn{1}{c|}{0.430} & 0.453 \\ \hline
VL-Rethinker-7B       & \multicolumn{1}{c|}{0.792} & 0.819 & \multicolumn{1}{c|}{0.418} & 0.439 \\ \hline
\end{tabular}
}
\label{tab:vector-space}
\end{table}

\subsection{Ablation Studies: Thinking vs Answer tokens. }

Table~\ref{tab:openvlthinker_tokens} compares the performance of OpenVLThinker across the full output of the model, thinking tokens, and answer tokens using multiple lexical and semantic similarity metrics. Overall, \textit{thinking tokens} consistently yield the highest semantic similarity scores, with the best BERTScore (0.872 Avg., 0.877 Max.) and Sentence-BERT cosine similarity (0.810 Avg., 0.834 Max.). In contrast, \textit{answer tokens} achieve relatively lower values across most metrics, particularly in ROUGE-1 (0.433 Avg.) and TF-IDF (0.671 Avg.), though they perform competitively in overlap coefficient (0.585 Avg., 0.647 Max.). The \textit{full output} generally provides balanced performance between the two extremes, showing stable lexical scores (e.g., Jaccard 0.285 Avg., 0.300 Max.) and strong TF-IDF cosine similarity (0.399 Avg., 0.421 Max.). These results highlight that reasoning-oriented thinking tokens capture richer semantic alignment, whereas concise answer tokens excel more in exact lexical overlap.

\begin{table}[h]
\caption{Thinking Tokens vs Answer Tokens of OpenVLThinker Model}
\resizebox{\linewidth}{!}{
\begin{tabular}{|c|cc|cc|cc|}
\hline
\multirow{2}{*}{Metric}                                                    & \multicolumn{2}{c|}{Full Output}   & \multicolumn{2}{c|}{Thinking Tokens} & \multicolumn{2}{c|}{Answer Tokens} \\ \cline{2-7} 
                                                                           & \multicolumn{1}{c|}{Avg.}  & Max.  & \multicolumn{1}{c|}{Avg.}   & Max.   & \multicolumn{1}{c|}{Avg.}  & Max.  \\ \hline
BERTScore                                                                  & \multicolumn{1}{c|}{0.864} & 0.869 & \multicolumn{1}{c|}{0.872}  & 0.877  & \multicolumn{1}{c|}{0.847} & 0.860 \\ \hline
\begin{tabular}[c]{@{}c@{}}Sentence-BERT \\ Cosine Similarity\end{tabular} & \multicolumn{1}{c|}{0.807} & 0.832 & \multicolumn{1}{c|}{0.810}  & 0.834  & \multicolumn{1}{c|}{0.772} & 0.808 \\ \hline
ROUGE-1                                                                    & \multicolumn{1}{c|}{0.527} & 0.559 & \multicolumn{1}{c|}{0.497}  & 0.543  & \multicolumn{1}{c|}{0.433} & 0.476 \\ \hline
ROUGE-L                                                                    & \multicolumn{1}{c|}{0.262} & 0.274 & \multicolumn{1}{c|}{0.260}  & 0.277  & \multicolumn{1}{c|}{0.248} & 0.277 \\ \hline
Jaccard Index                                                              & \multicolumn{1}{c|}{0.285} & 0.300 & \multicolumn{1}{c|}{0.283}  & 0.304  & \multicolumn{1}{c|}{0.227} & 0.253 \\ \hline
Overlap Coefficient                                                        & \multicolumn{1}{c|}{0.525} & 0.568 & \multicolumn{1}{c|}{0.511}  & 0.542  & \multicolumn{1}{c|}{0.585} & 0.647 \\ \hline
TF-IDF                                                                     & \multicolumn{1}{c|}{0.795} & 0.820 & \multicolumn{1}{c|}{0.763}  & 0.798  & \multicolumn{1}{c|}{0.671} & 0.699 \\ \hline
TF-IDF Cosine Similarity                                                   & \multicolumn{1}{c|}{0.399} & 0.421 & \multicolumn{1}{c|}{0.395}  & 0.418  & \multicolumn{1}{c|}{0.340} & 0.367 \\ \hline
\end{tabular}
}
\label{tab:openvlthinker_tokens}
\end{table}

\section{Conclusion and Future Work} \label{sec:conclusion}

In this paper, we present the ImageNet-Think-250K dataset, a large-scale reasoning tokens from state-of-the-art Vision Language Models. Our dataset offers explicit multimodal reasoning processes to enable the development of more interpretable and robust reasoning VLMs. The main contributions of this work include the creation of the largest publicly available dataset with explicit multimodal reasoning annotations, comprising 250,000 images with 500,000 thinking-answer pairs and benchmarking them across five open source reasoning VLMs. 

\textbf{Future Work:} In future, we plan to expand the dataset to include additional image modalities (video, audio) and incorporate more recent thinking models as they become available. We also plan to include multi-lingual reasoning tokens and domain-specific applications. We also plan to develop more sophisticated metrics for assessing reasoning quality and output accuracy. Leveraging ImageNet-Think to develop novel architectures specifically designed for explicit reasoning generation, including hierarchical reasoning models, multi-step verification systems, and interactive reasoning frameworks.







\section*{Acknowledgements}
This research used resources of the Argonne Leadership Computing Facility, a U.S. Department of Energy (DOE) Office of Science user facility at Argonne National Laboratory and is based on research supported by the U.S. DOE Office of Science-Advanced Scientific Computing Research Program, under Contract No. DE-AC02-06CH11357

{
    \small
    \bibliographystyle{ieeenat_fullname}
    \bibliography{main}
}

\end{document}